\documentclass[conference]{IEEEtran}
\ifCLASSINFOpdf
\else
\fi
\usepackage{float}
\usepackage{graphicx}
\usepackage{booktabs}
\usepackage[colorlinks=black,linkcolor=black,hyperindex,CJKbookmarks]{hyperref}
\usepackage{CJK}
\usepackage{indentfirst}
\usepackage{amsmath}
\usepackage{amsfonts}    
\begin{document}
\title{A Deep Graph Embedding Network Model \\for Face Recognition}
\author{\IEEEauthorblockN{Yufei Gan, Teng Yang, Chu He}
\IEEEauthorblockA{Electronic  Information  School,  Wuhan  University, Wuhan 430072, China\\ 
Email: ganyufei@whu.edu.cn, tengyang@whu.edu.cn, chuhe@whu.edu.cn}
}

\maketitle

\begin{abstract}
In this paper, we propose a new deep learning network ``GENet'', it combines the multi-layer network architecture and graph embedding framework. Firstly, we use simplest unsupervised learning PCA/LDA as first layer to generate the low-level feature. Secondly, many cascaded dimensionality reduction layers based on graph embedding framework are applied to GENet. Finally, a linear SVM classifier is used to classify dimension-reduced features. The experiments indicate that higher classification accuracy can be obtained by this algorithm on the CMU-PIE, ORL, Extended Yale B dataset.
\end{abstract}

\IEEEpeerreviewmaketitle

\section{Introduction}
The task of image classification is fundamental for many computer vision tasks. Currently, many approaches have been proposed to solve this task, and many researches about image classification are related to dimensionality reduction.

In order to avoid the curse of dimensionality issue, many dimensionality reduction algorithms have been proposed, and Yan \cite{yan2007graph}\cite{yan2005graph} has proposed a common framework based on the direct graph embedding to unify these algorithms. In the framework of graph embedding, we can use a unified view for understanding and explaining many of the popular dimensionality reduction algorithms, and a new dimensionality reduction algorithm called Marginal Fisher Analysis (MFA) has been proposed.

Recent research \cite{chan2014pcanet}\cite{bengio2009learning} shows the advantage of deep network in image classification, However, the framework of graph embedding do not have a multi-layer construction. A multi-layer construction can contribute the performance the of the single network. 

GENet use deep network to enhance the performance of the framework of graph embedding. In GENet, a cascaded dimensionality reduction algorithm based on graph embedding network is applied to reduce feature dimensions.

Deep learning network mostly use an unsupervised learning as first layer and these unsupervised learning actually learn the feature from the data. The framework of graph embedding also learns the mapping that transforms high-dimensional representation into the desired low-dimensional representation.

Our contribution is that we proposed a simple and fast deep network for comparing and justifying other more advanced deep learning components or architectures like CNNs \cite{krizhevsky2012imagenet}. Once the parameters are fixed, training GENet is extremely simple and efficient, for the filter learning in GENet does not involve regularized parameters and does not require numerical optimization solver. 

As the research further develops, researchers find the fact that the convolutional deep neural network (CNNs) has weak classification capacity in high-level layer \cite{sharif2014cnn} when compared to SVM. So SVM has been applied to replace the high-level layer recently, and our GENet also uses SVM as classifier.

\section{Graph Embedding Framework}
\begin{figure*}[!bpht]
\centering
\includegraphics[width=7in]{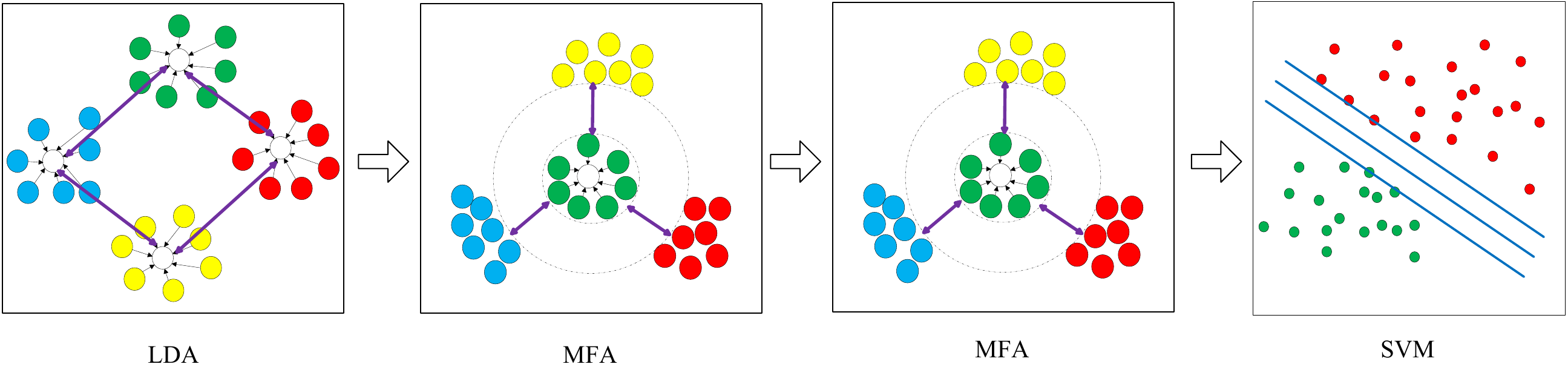}
\caption{The construction of GENet, three cascaded dimensionality reduction algorithms based on the framework of graph embedding are shown in this picture. Different color dots represent different classes, hollow dots in the center of each kind of class represent the clustering center. We use unsupervised learning algorithms as first layer (for instance, in this figure, we select LDA). and two layers MFA as supervised learning layer. The classifier is SVM to efficiently classify the low-dimensional feature.} \label{figure:construction}
\end{figure*}
To represent each vertex of a graph as a low-dimensional vector, Yan \cite{yan2007graph} has proposed a general framework called graph embedding to offer a unified view for understanding and explaining many of the popular dimensionality reduction algorithms. In graph embedding,  we use graph to describe the manifold structure of data, and we denote the sample set as $X=\left[x_1,x_2,\cdots,x_N\right],x_i\in\mathbb{R}^m$. In the supervised learning problem, the $N_c$ class labels are assumed as $c_i\in\{1,2,\cdots,N_c\}$ and denote $\pi$ as the index set of samples belonging to class $c$. For an undirected weighted graph $G=\{X,W\}$ with vertex set $X$ and similarity matrix $W\in\mathbb{R}^{N\times{N}}$. And in matrix $W$, each element of the real symmetric matrix $W$ measures, for a pair of vertices, its similarity, which may be negative.

The diagonal matrix $D$ and the Laplacian matrix $L$ of a graph $G$ are defined as
\begin{equation}\label{laplace}
L=D-W,~D_{ii}=\sum_{j\neq{i}}W_{ij},~\forall{i}.
\end{equation}

We define an intrinsic graph to be the graph $G$ itself and a penalty graph $G^p=\{X,W^p\}$ as a graph whose vertices $X$ are the same as those of $G$, but whose edge weight matrix $W^p$ corresponds to the similarity characteristics that are to be suppressed in the dimension-reduced feature space. 
we represent the low-dimensional representations of the vertices as a vector $y=\left[y_1,y_2,\cdots,y_N\right]^T$, where $y_i$ is the low-dimensional representation of vertex $x_i$.

And for a dimensionality reduction problem, we require an intrinsic graph $G$ and, optionally, a penalty graph $G^p$ as input. Our graph-preserving criterion is given as follows:
\begin{equation}\label{y*}
y^*=\arg\min_{y^TBy=d}\sum_{i\neq{j}}\|y_i-y_j\|^{2}W_{ij}=\arg\min_{y^TBy=d}y^TLy,
\end{equation}
where $d$ is a constant and $B$ is the constraint matrix ($B=L^p=D^p-W^p$)

In the framework of graph embedding, we can describe the traditional dimensionality reduction algorithms in the same framework by setting intrinsic graph $G$ and penalty graph $G^p$. 

PCA \cite{jolliffe2005principal} finds and removes the projection directions with maximal variance.To put it in another word, it finds and removes the projection direction with minimal variance:
\begin{equation}
\begin{aligned}
&w^*&=&\arg\min_{w^Tw=1}w^TCw~~~with&&&&&&&&&&&&&&& \\
&C&=&\frac{1}{N}\sum^N_{i=1}(x_i-\bar{x})(x_i-\bar{x})^T&& \\
&&=&\frac{1}{N}X(I-\frac{1}{N}ee^T)X^T&
\end{aligned}
\end{equation}
where $\bar{x}$ is the mean of all samples and $e$ is a $N$ dimensional vector with $e=\left[1,1,\cdots,1\right]^T$.

LDA \cite{martinez2001pca} searches for the directions that are most effective for discrimination by minimizing the ratio between the intraclass and interclass scatters:
\begin{equation}
\begin{aligned}
&w^*&=&\arg\min_{w^T S_B w=d}w^T S_W w=\arg\min_w \frac{w^T S_W w}{w^T S_B w}&\\
&&=&\arg\min_w \frac{w^T S_W w}{w^T C w}&\\
&S_W&=&\sum^N_{i=1}(x_i-\bar{x}^{c_i})(x_i-\bar{x}^{c_i})^T& \\
&&=&X(I-\sum^{N_c}_{c=1}\frac{1}{n_c}e^{c}e^{cT})X^T  &\\
&S_B&=&\sum^{N_c}_{c=1}n_c(\bar{x}^c-\bar{x})(\bar{x}^c-\bar{x})^T& \\
&&=&NC-S_W&
\end{aligned}
\end{equation}
where $\bar{x}^c$ is the mean of the $c$-th class, and $e^c$ is an $N$ dimensional vector with $e^c(i)=1,~if~c=l_i;0,otherwise$.

In MFA (Marginal Fisher Analysis), the intrinsic graph characterizes the intraclass compactness and connects each data point with its neighboring points of the same class, while the penalty graph connects the marginal points and characterizes the interclass separability. 
\begin{equation}
\begin{aligned}
&w^*&=&\arg\min_w \frac{w^T X(D-W)X^T w}{w^T X(D^p-W^p)X^T w}&\\
&\tilde{S}_t&=&\sum_i\sum_{i\in N^+_{k_1}(j)~or~j\in N^+_{k_1}(i)}\|w^T x_i -w^T x_j\|^2& \\
&&=&2w^T X(D-W)X^Tw,& \\
&&&W_{i,j}=\left\{
\begin{aligned}
&1~,~if~i\in N^+_{k_1}(j)~or~j\in N^+_{k_1}(i)&\\
&0~,~else.&
\end{aligned}
\right.&\\
&&&D^w_{ii}=\sum_j{W^w_{ij}}.&\\
&\tilde{S}_p&=&\sum_i\sum_{(i,j)\in P_{k_2}(c_i)~or~j\in P_{k_2}(c_j)}\|w^T x_i -w^T x_j\|^2& \\
&&=&2w^T X(D^P-W^P)X^Tw,& \\
&&&W^p_{i,j}=\left\{
\begin{aligned}
&1~,~if~(i,j)\in P_{k_2}(c_i)~or~(i,j)\in P_{k_2}(c_j)&\\
&0~,~else.&
\end{aligned}
\right.&\\
&&&D^p_{ii}=\sum_j{W^p_{ij}}&
\end{aligned}
\end{equation}
where $N^+_{k_1}(i)$ indicates the index set of the $k_1$ nearest neighbors of the sample $x_i$ in the class, and $P_{k_1}(c)$ is a set of data pairs that are the $k_2$ nearest pairs among the set $\{(i,j),i\in\pi_c,j\notin\pi_c\}$.

\begin{table}[!ht]
\caption{}\label{tableGE}
\centering
The Common Graph Embedding View \\
for the Most Popular Dimensionality Reduction Algorithms

~\\
\centering
\begin{tabular}{cc} \toprule
Algorithm				&	W\&B Definition  	\\ \midrule
PCA/KPCA/2DPCA		& $W_{ij}=\frac{1}{N},i\neq{j};B=I$\\  \midrule
LDA/KDA/2DLDA/DATER	&$W_{ij}=\delta_{c_i,c_j}/n_c;B=I-\frac{1}{N}ee^T$\\ \bottomrule
\end{tabular}
\end{table}

\section{GENet}
In the Graph Embedding, some dimensionality reduction algorithms can be applied to our GENet, such as PCA, LDA, IOSMAP, LLE, and so on. 
\begin{figure*}[!bpht]
\centering
\includegraphics[scale=0.5]{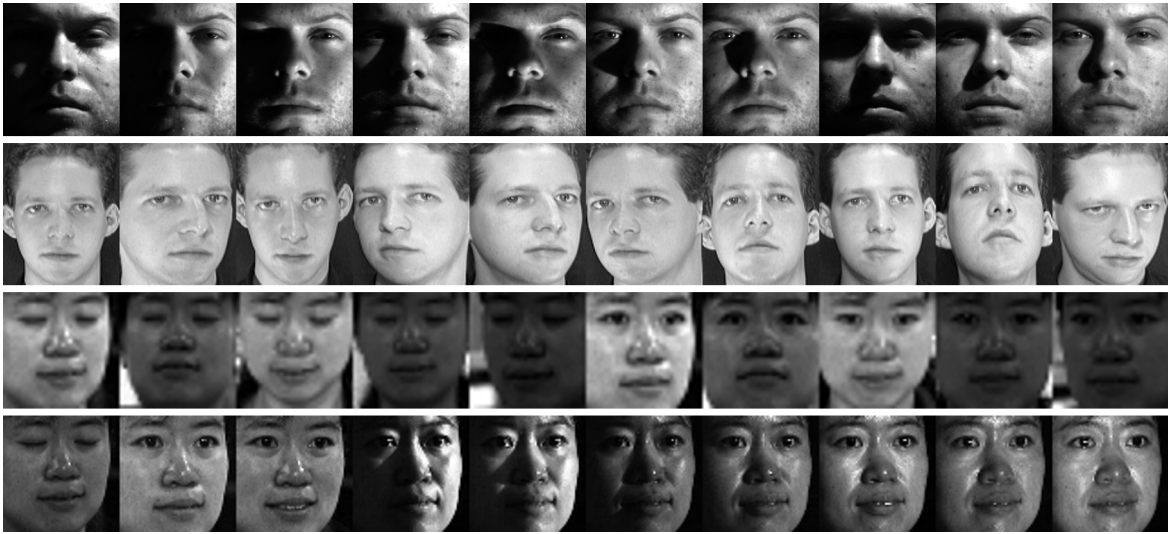}
\caption{The sample images cropped from the face database Extended Yale B (first row), ORL (second row), PIE\_32x32 (third row), and Pose05\_64x64 (fourth row), respectively.}
\end{figure*}
In order to research the advantage of the deep network construction, we simply use the simplest dimensionality reduction linear algorithms--PCA and LDA. Moreover, in consideration of that Marginal Fisher Analysis show an competitive performance in Yan's \cite{yan2007graph} paper, we also introduce liner MFA to our GENet.

In our GENet, we use unsupervised learning algorithm as the first layer to obtain sufficient low-level-feature. In the experiment section, we will find the fact that use unsupervised learning algorithm as the first layer perform better than supervised learning algorithm.

The construction of GENet is shown in Figure \ref{figure:construction}. In the figure, three cascaded dimensionality reduction algorithms based on the framework of graph embedding are shown in this picture. Different color dots represent different classes, hollow dots in the center of each kind of class represent the clustering center.

\section{Experiment}
In our experiments, we use the Extended Yale Face Database B, CMU PIE, and ORL databases for face recognition to evaluate our GENet.

With the following reasons, we decide to use the dengcai versions \cite{CHHH07}, \cite{CHH07b}, \cite{CHHZ06}, \cite{HYHNZ05} of Extended Yale Face Database B, CMU PIE and ORL face sets:

A. Faces are already standardized according to eye locations -- so that when we compare the performances of identification algorithms, we do not need to worry if the standardization approaches are the same. 

B. There are two versions, 32$\times$32 pixels and 64$\times$64 pixels, available for each set, so that we can see if our approach works for different sizes.

C. The above UIUC site provides many holistic algorithms, including newly developed ones, in source codes, they also provided best results of many algorithms (5 to 10 algorithms) for 32$\times$32 versions of above data sets.

In the dengcai versions of Extended Yale Face Database B, PIE and ORL face sets, the Extended Yale Face Database B has 38 individuals and around 64 near frontal images under different illuminations per individual, the data file -- YaleB\_32x32 contains 2,414 images which are cropped and resize to 32x32 pixels. And CMU PIE contains 41,368 images of 68 people, each person under 13 different poses, 43 different illumination conditions, and with 4 different expressions, the CMU PIE data file -- PIE\_32x32 contains 11,554 images which are cropped and resize to 32x32 pixels. The ORL face sets have ten different images of each of 40 distinct subjects. For some subjects, the images were taken at different times, varying the lighting, facial expressions (open / closed eyes, smiling / not smiling) and facial details (glasses / no glasses). All the images were taken against a dark homogeneous background with the subjects in an upright, frontal position (with tolerance for some side movement), and the data file -- ORL\_32x32 contains 400 images which are cropped and resize to 32x32 pixels.

The implementation of the PCA/LDA/MFA are obtained from dengcai's personal homepage: \url{http://www.cad.zju.edu.cn/home/dengcai/Data/DimensionReduction.html}

The experiment on ORL use five different training data size (from 1 to 5) to train GENet, and the remaining images are used for testing. However in PIE dataset and Extended Yale B dataset, the number of images of each person are different, so use the same test data size and the remaining images are used for training.

We use PCA+MFA+PCA+MFA (100, 70, 60, 40), PCA+MFA (100, 40), LDA+MFA+LDA+MFA (100, 70, 60, 40), LDA+MFA (100, 40), LDA+MFA+PCA+MFA (100, 70, 60, 40), PCA+MFA+LDA+MFA (100, 70, 60, 40), PCA+MFA+MFA (100, 70, 40), LDA+MFA+MFA (100, 70, 40)  these combinations, the the numbers in parentheses are the corresponding feature dimensions, and the parameters of MFA are set to $k_1=10, k_2=500$ in the experiment on ORL,  $k_1=2, k_2=440$ in the experiment on PIE, and $k_1=10, k_2=500$ in the experiment on Extended Yale B. The results are shown in Table \ref{table:ORL}, Table \ref{table:PIE} and Table \ref{table:Extended Yale B}
 
\begin{table}[!ht]
\caption{}\label{table:ORL}
\centering
Face Recognition Accuracies of PCA+MFA+PCA+MFA , PCA+MFA, LDA+MFA+LDA+MFA, LDA+MFA, LDA+MFA+PCA+MFA, PCA+MFA+LDA+MFA, PCA+MFA+MFA, LDA+MFA+MFA on the ORL Database. (numbers in parentheses are the corresponding feature dimensions)

~\\
\centering
\begin{tabular}{cccccc} \toprule
Train/Test				&	1/9	&	2/8	&	3/7 	& 	4/6 	& 	5/5	\\ \midrule
PCA+MFA+PCA+MFA	& 65.00\%	&80.00\%	& 75.00\%	& 89.16\%	&	96.50\%\\  \midrule
PCA+MFA \cite{yan2007graph}			& 73.61\%&80.31\%	& 83.21\%	& 88.75\%&	94.00\%\\  \midrule
LDA+MFA+LDA+MFA	& 70.27\%	&77.81\%	& 72.85\%	& 72.50\%	&	69.00\%\\  \midrule
LDA+MFA				& 64.72\%	&79.06\%	& 72.14\%	& 68.75\%	&	76.50\%\\  \midrule
LDA+MFA+PCA+MFA	& 65.55\%	&74.37\%	& 79.28\%	& 72.91\%	&	70.00\%\\  \midrule
PCA+MFA+LDA+MFA	& 67.77\%	&77.18\%	& 80.00\%	& \textbf{92.91}\%	&	70.00\%\\  \midrule
PCA+MFA+MFA		& \textbf{74.16}\%	&\textbf{80.62}\%	& \textbf{88.21}\%	& 88.33\%	&	\textbf{97.50}\%\\  \midrule
LDA+MFA+MFA		& 67.50\%	&75.93\%	& 71.78\%	& 69.16\%	&	95.50\%\\  \bottomrule
\end{tabular}
~\\
~\\
Note that the approach PCA+MFA is proposed in \cite{yan2007graph}.
\end{table}

\begin{table}[!ht]\
\caption{}\label{table:PIE}
\centering
Face Recognition Accuracies of PCA+MFA+PCA+MFA , PCA+MFA, LDA+MFA+LDA+MFA, LDA+MFA, LDA+MFA+PCA+MFA, PCA+MFA+LDA+MFA, PCA+MFA+MFA, LDA+MFA+MFA on the Pose05\_64x64 Database. (numbers in parentheses are the corresponding feature dimensions)

~\\
\centering
\begin{tabular}{cccccc} \toprule
Test					&30		&	20	& 	10		\\  \midrule
PCA+MFA+PCA+MFA	&\textbf{79.60}\%	&\textbf{84.92}\%	&	\textbf{91.61}\%		\\  \midrule
PCA+MFA \cite{yan2007graph}				&54.26\%	&82.35\%	&	80.88\%		\\  \midrule
LDA+MFA+LDA+MFA	&2.64\%	&1.47\%	&	6.02\%	\\  \midrule
LDA+MFA				&2.64\%	&1.47\%	&	6.02\%	\\  \midrule
LDA+MFA+PCA+MFA	&2.74\%	&3.01\%	&	6.02\%	\\  \midrule
PCA+MFA+LDA+MFA	&72.20\%	&79.55\%	&	82.35\%	\\  \midrule
PCA+MFA+MFA		&59.01\%	& 81.25\%	&	90.29\%	\\  \midrule
LDA+MFA+MFA		&0.78\%	& 2.79\%	&	2.35\%	\\  \bottomrule
\end{tabular}
~\\
~\\
Note that the approach PCA+MFA is proposed in \cite{yan2007graph}.
\end{table}
\begin{table}[!ht]\
\caption{}\label{table:Extended Yale B}
\centering
Face Recognition Accuracies of PCA+MFA+PCA+MFA , PCA+MFA, LDA+MFA+LDA+MFA, LDA+MFA, LDA+MFA+PCA+MFA, PCA+MFA+LDA+MFA, PCA+MFA+MFA, LDA+MFA+MFA on the YaleB\_32x32 Database. (numbers in parentheses are the corresponding feature dimensions, and the number of test images is 1)

~\\
\centering
\begin{tabular}{cccccc} \toprule
PCA+MFA+PCA+MFA		&	61.05\%	&PCA+MFA \cite{yan2007graph}		&	73.68\%	\\  \midrule
LDA+MFA+LDA+MFA		&	82.10\%	&LDA+MFA		&	82.10\%\\  \midrule
LDA+MFA+PCA+MFA		&	76.84\%	&PCA+MFA+LDA+MFA		&	73.68\%\\  \midrule
PCA+MFA+MFA			&	62.10\%	&LDA+MFA+MFA			&	\textbf{84.21}\%	\\  \bottomrule
\end{tabular}
~\\
~\\
Note that the approach PCA+MFA is proposed in \cite{yan2007graph}.
\end{table}
Comparing the experiment on Extended Yale B with the experiment on PIE, We can know that LDA algorithm have not a good performance on PIE dataset, but it have a good performance on Extended Yale B dataset, the results may due to limitations of data distribution assumptions.

Another observation is a obvious result -- in most cases, the GENet with multi-layer performance better than GENet with single-layer.

\section{Conclusion}
In this paper, we proposed a deep learning network based on the graph embedding. GENet combine the deep network and the framework of graph embedding, it uses a multi-layer graph embedding construction to classify images. Using GENet to compute filters does not require numerical optimization solver so the training process can be extremely efficient. Because of the of framework of the graph embedding, GENet can be treated as the cascaded network, the different of each layer is the setting of intrinsic graph and penalty graph.

Our results indicate that GENet can perform fast and accuracy in face recognition datasets such as Extended Yale B, CMU PIE, and ORL databases. Moreover, the results show the fact that multi-layer construction can perform better than single-layer construction. GENet makes it possible to find a adaptive algorithm to reduce dimensions and try to avoid the impact of data distribution assumptions.

In feature work, we hope to apply the extensions of graph embedding -- kernel and tensor to our GENet, and we hope to find the effective method to let the parameters of GENet, specially the algorithm of each layer, can learn from the data.

\section*{Acknowledgement}
The work was supported by the National Key Basic Research and Development Program of China (973 Program) (No.2013CB733404), NSFC grant (No.41371342, No.61331016) and the China Postdoctoral Science Foundation funded project and the Natural Science Foundation of Hubei Province.

\bibliographystyle{IEEEtran}
\bibliography{G}

\begin{thebibliography}{10}
\providecommand{\url}[1]{#1}
\csname url@samestyle\endcsname
\providecommand{\newblock}{\relax}
\providecommand{\bibinfo}[2]{#2}
\providecommand{\BIBentrySTDinterwordspacing}{\spaceskip=0pt\relax}
\providecommand{\BIBentryALTinterwordstretchfactor}{4}
\providecommand{\BIBentryALTinterwordspacing}{\spaceskip=\fontdimen2\font plus
\BIBentryALTinterwordstretchfactor\fontdimen3\font minus
  \fontdimen4\font\relax}
\providecommand{\BIBforeignlanguage}[2]{{%
\expandafter\ifx\csname l@#1\endcsname\relax
\typeout{** WARNING: IEEEtran.bst: No hyphenation pattern has been}%
\typeout{** loaded for the language `#1'. Using the pattern for}%
\typeout{** the default language instead.}%
\else
\language=\csname l@#1\endcsname
\fi
#2}}
\providecommand{\BIBdecl}{\relax}
\BIBdecl

\bibitem{yan2007graph}
S.~Yan, D.~Xu, B.~Zhang, H.-J. Zhang, Q.~Yang, and S.~Lin, ``Graph embedding
  and extensions: a general framework for dimensionality reduction,''
  \emph{Pattern Analysis and Machine Intelligence, IEEE Transactions on},
  vol.~29, no.~1, pp. 40--51, 2007.

\bibitem{yan2005graph}
S.~Yan, D.~Xu, B.~Zhang, and H.-J. Zhang, ``Graph embedding: A general
  framework for dimensionality reduction,'' in \emph{Computer Vision and
  Pattern Recognition, 2005. CVPR 2005. IEEE Computer Society Conference on},
  vol.~2.\hskip 1em plus 0.5em minus 0.4em\relax IEEE, 2005, pp. 830--837.

\bibitem{chan2014pcanet}
T.-H. Chan, K.~Jia, S.~Gao, J.~Lu, Z.~Zeng, and Y.~Ma, ``Pcanet: A simple deep
  learning baseline for image classification?'' \emph{arXiv preprint
  arXiv:1404.3606}, 2014.

\bibitem{bengio2009learning}
Y.~Bengio, ``Learning deep architectures for ai,'' \emph{Foundations and
  trends{¥textregistered} in Machine Learning}, vol.~2, no.~1, pp. 1--127,
  2009.

\bibitem{krizhevsky2012imagenet}
A.~Krizhevsky, I.~Sutskever, and G.~E. Hinton, ``Imagenet classification with
  deep convolutional neural networks,'' in \emph{Advances in neural information
  processing systems}, 2012, pp. 1097--1105.

\bibitem{sharif2014cnn}
A.~Sharif~Razavian, H.~Azizpour, J.~Sullivan, and S.~Carlsson, ``Cnn features
  off-the-shelf: an astounding baseline for recognition,'' \emph{arXiv preprint
  arXiv:1403.6382}, 2014.

\bibitem{jolliffe2005principal}
I.~Jolliffe, \emph{Principal component analysis}.\hskip 1em plus 0.5em minus
  0.4em\relax Wiley Online Library, 2005.

\bibitem{martinez2001pca}
A.~M. Mart{¥'¥i}nez and A.~C. Kak, ``Pca versus lda,'' \emph{Pattern Analysis
  and Machine Intelligence, IEEE Transactions on}, vol.~23, no.~2, pp.
  228--233, 2001.

\bibitem{CHHH07}
D.~Cai, X.~He, Y.~Hu, J.~Han, and T.~Huang, ``Learning a spatially smooth
  subspace for face recognition,'' in \emph{Proc. IEEE Conf. Computer Vision
  and Pattern Recognition Machine Learning (CVPR'07)}, 2007.

\bibitem{CHH07b}
D.~Cai, X.~He, and J.~Han, ``Spectral regression for efficient regularized
  subspace learning,'' in \emph{Proc. Int. Conf. Computer Vision (ICCV'07)},
  2007.

\bibitem{CHHZ06}
D.~Cai, X.~He, J.~Han, and H.-J. Zhang, ``Orthogonal laplacianfaces for face
  recognition,'' \emph{IEEE Transactions on Image Processing}, vol.~15, no.~11,
  pp. 3608--3614, 2006.

\bibitem{HYHNZ05}
X.~He, S.~Yan, Y.~Hu, P.~Niyogi, and H.-J. Zhang, ``Face recognition using
  laplacianfaces,'' \emph{IEEE Trans. Pattern Anal. Mach. Intelligence},
  vol.~27, no.~3, pp. 328--340, 2005.

\end{thebibliography}

\end{document}